\documentclass[conference]{IEEEtran}
\IEEEoverridecommandlockouts                              %

\usepackage{algorithm}
\usepackage{algpseudocode}
\usepackage{kotex}
\usepackage{xcolor}
\usepackage{booktabs}
\usepackage{multirow}
\usepackage{xspace}
\usepackage{microtype}
\usepackage{longtable}

\usepackage{cite}
\usepackage{amsmath}
\usepackage{amssymb}
\usepackage{amsfonts}

\usepackage[final]{graphicx}
\usepackage{textcomp}
\usepackage{etoolbox}
\usepackage{comment}
\usepackage{hyperref}
\usepackage{cleveref}
\usepackage{makecell}
\usepackage{flushend}
\usepackage{subcaption}

\title{
Learning Nonverbal Cues in Multiparty Social Interactions for Robotic Facilitators}
\begin{document}

\author{\IEEEauthorblockN{Antonio Martin-Ozimek}
\IEEEauthorblockA{
\textit{Honda Research Institute}\\
Wako, Japan \\
\textit{University of Alberta}\\
Edmonton, Canada \\
antonio2@ualberta.ca}%
\and
\IEEEauthorblockN{Isuru Jayarathne}
\IEEEauthorblockA{
\textit{Honda Research Institute}\\
Wako, Japan \\
isuru.jayarathne@jp.honda-ri.com}%
\and
\IEEEauthorblockN{Su Larb Mon}
\IEEEauthorblockA{
\textit{Honda Research Institute}\\
Wako, Japan \\
su.larbmon@jp.honda-ri.com}%
\and
\IEEEauthorblockN{Jouhyeong Chew}
\IEEEauthorblockA{
\textit{Honda Research Institute}\\
Wako, Japan \\
jouhyeong.chew@jp.honda-ri.com}%
}%
\maketitle

\begin{abstract}
Conventional behavior cloning (BC) models often struggle to replicate the subtleties of human actions. Previous studies have attempted to address this issue through the development of a new BC technique: Implicit Behavior Cloning (IBC). This new technique consistently outperformed the conventional Mean Squared Error (MSE) BC models in a variety of tasks. Our goal is to replicate the performance of the IBC model by Florence [in Proceedings of the 5th Conference on Robot Learning, 164:158-168, 2022], for social interaction tasks using our custom dataset. While previous studies have explored the use of large language models (LLMs) for enhancing group conversations, they often overlook the significance of non-verbal cues, which constitute a substantial part of human communication. We propose using IBC to replicate nonverbal cues like gaze behaviors. The model is evaluated against various types of facilitator data and compared to an explicit, MSE BC model. Results show that the IBC model outperforms the MSE BC model across session types using the same metrics used in the previous IBC paper. Despite some metrics showing mixed results which are explainable for the custom dataset for social interaction, we successfully replicated the IBC model to generate nonverbal cues.  
Our contributions are (1) the replication and extension of the IBC model, and (2) a nonverbal cues generation model for social interaction. These advancements facilitate the integration of robots into the complex interactions between robots and humans, e.g., in the absence of a human facilitator.

\end{abstract}

\newcommand{\todo}[1]{\textcolor{blue}{{TODO: #1}}}
\newcommand{\taery}[1]{\textcolor{violet}{{Taery: #1}}}
\newcommand{\sehoon}[1]{\textcolor{red}{{Sehoon: #1}}} 
\newcommand{\wenhao}[1]{\textcolor{blue}{{Wenhao: #1}}} 
\newcommand{\greg}[1]{\textcolor{cyan}{{Greg: #1}}}

\newcommand{\newtext}[1]{#1}
\newcommand{\original}[1]{\textcolor{magenta}{Original: #1}}
\newcommand{\eqnref}[1]{Equation~(\ref{eq:#1})}
\newcommand{\figref}[1]{Figure~\ref{fig:#1}}
\renewcommand{\algref}[1]{Algorithm~\ref{alg:#1}}
\newcommand{\tabref}[1]{Table~\ref{tab:#1}}
\newcommand{\secref}[1]{Section~\ref{sec:#1}}
\newcommand{\mypara}[1]{\noindent\textbf{{#1}.}}

\long\def\ignorethis#1{}

\newcommand{\etal}{{\em{et~al.}\ }}
\newcommand{\eg}{e.g.\ }
\newcommand{\ie}{i.e.\ }

\newcommand{\figtodo}[1]{\framebox[0.8\columnwidth]{\rule{0pt}{1in}#1}}



\newcommand{\pdd}[3]{\ensuremath{\frac{\partial^2{#1}}{\partial{#2}\,\partial{#3}}}}

\newcommand{\mat}[1]{\ensuremath{\mathbf{#1}}}
\newcommand{\set}[1]{\ensuremath{\mathcal{#1}}}

\newcommand{\vc}[1]{\ensuremath{\mathbf{#1}}}
\newcommand{\vEndEff}{\ensuremath{\vc{d}}}
\newcommand{\vRelMove}{\ensuremath{\vc{r}}}
\newcommand{\sSet}{\ensuremath{S}}

\newcommand{\vControl}{\ensuremath{\vc{u}}}
\newcommand{\vPoint}{\ensuremath{\vc{p}}}
\newcommand{\sSpringCoef}{{\ensuremath{k_{s}}}}
\newcommand{\sDamperCoef}{{\ensuremath{k_{d}}}}
\newcommand{\vHandle}{\ensuremath{\vc{h}}}
\newcommand{\vForce}{\ensuremath{\vc{f}}}

\newcommand{\mTransChain}{\ensuremath{\vc{W}}}
\newcommand{\mRotateTrans}{\ensuremath{\vc{R}}}
\newcommand{\sJoint}{\ensuremath{q}}
\newcommand{\vJoint}{\ensuremath{\vc{q}}}
\newcommand{\mJoint}{\ensuremath{\vc{Q}}}
\newcommand{\mMass}{\ensuremath{\vc{M}}}
\newcommand{\sMass}{\ensuremath{{m}}}
\newcommand{\vGravity}{\ensuremath{\vc{g}}}
\newcommand{\vConstr}{\ensuremath{\vc{C}}}
\newcommand{\sConstr}{\ensuremath{C}}
\newcommand{\vCOM}{\ensuremath{\vc{x}}}
\newcommand{\sGeneralForce}[1]{\ensuremath{Q_{#1}}}
\newcommand{\vStateVar}{\ensuremath{\vc{y}}}
\newcommand{\vControlVar}{\ensuremath{\vc{u}}}
\newcommand{\tr}[1]{\ensuremath{\mathrm{tr}\left(#1\right)}}

%
%

\renewcommand{\choose}[2]{\ensuremath{\left(\begin{array}{c} #1 \\ #2 \end{array} \right )}}

\newcommand{\gauss}[3]{\ensuremath{\mathcal{N}(#1 | #2 ; #3)}}

\newcommand{\pctab}{\hspace{0.2in}}
\newenvironment{pseudocode} {\begin{center} \begin{minipage}{\textwidth}
                             \normalsize \vspace{-2\baselineskip} \begin{tabbing}
                             \pctab \= \pctab \= \pctab \= \pctab \=
                             \pctab \= \pctab \= \pctab \= \pctab \= \\}
                            {\end{tabbing} \vspace{-2\baselineskip}
                             \end{minipage} \end{center}}
\newenvironment{items}      {\begin{list}{$\bullet$}
                              {\setlength{\partopsep}{\parskip}
                                \setlength{\parsep}{\parskip}
                                \setlength{\topsep}{0pt}
                                \setlength{\itemsep}{0pt}
                                \settowidth{\labelwidth}{$\bullet$}
                                \setlength{\labelsep}{1ex}
                                \setlength{\leftmargin}{\labelwidth}
                                \addtolength{\leftmargin}{\labelsep}
                                }
                              }
                            {\end{list}}
\newcommand{\newfun}[3]{\noindent\vspace{0pt}\fbox{\begin{minipage}{3.3truein}\vspace{#1}~ {#3}~\vspace{12pt}\end{minipage}}\vspace{#2}}

\newcommand{\key}{\textbf}
\newcommand{\fun}{\textsc}



\section{Introduction}

Spurred on by the advent of large language models (LLM), our lives are becoming increasingly enmeshed with artificial intelligence~\cite{chen2024artificial}. There is a drive to integrate these machine learning models into increasingly complex situations~\cite{hickok2024public, chiang2024enhancing, zhang2022storybuddy}. This paper focuses on replicating the results of one of these models with naturalistic data from a social interaction scene, i.e., multiparty facilitation. Current studies have delved into the use of an LLM as a facilitator for group conversations~\cite{chiang2024enhancing, liu2024peergpt, agent2021agent}. However, utilizing only an LLM as a facilitator leads to a significant communication gap: non-verbal behavior, which constitutes a large portion of social interactions~\cite{phutela2015importance}. Ishi et al. imply that there is a correlation between one's gaze behavior in a conversation and the accompanying gaze behavior of the people with whom they are conversing~\cite{eyegaze2021}. Meaning that gaze alone can have a great effect on the course of conversations. Our research focuses on extending IBC to generate nonverbal cues like gaze and pose during multiparty facilitation, prioritizing the former. We also evaluated the limitations of IBC for these social interaction tasks. 
We are building on the work of S. Gillet et al.~\cite{gillet2022learning} who present their findings on using a robotic facilitator to balance engagement using gaze behavior in a social interaction between a teacher and a student. The robot facilitator is used to enhance language learning in a gamified learning scenario between a native speaker and a student. We hope to build on their idea by expanding it to more than just two participants. 

The work done by R. Rosenberg-Kima et al. is a great demonstration of a robot facilitator interacting with a somewhat large group~\cite{rosenberg2020}. Their paper seeks to improve learning interactions in a classroom where the robot facilitator takes on the role of a teaching assistant during lectures. We aim to extend their work by looking at similar information-sharing settings in large group interactions but from the point of view of the teacher instead of his assistant. We are aiming to replicate and generate the gaze behavior of the main facilitator in a large group social interaction setting.
While many options are available for behavior generation models, we are influenced by the work done by Florence et al.~\cite{pmlr-v164-florence22a} and S.Gillet et al. ~\cite{gillet2022learning}. Their research exemplifies the effectiveness of BC models in closely imitating human behaviors, demonstrating that such models can learn and replicate complex tasks with high fidelity by leveraging human demonstration data. Furthermore, traditional BC methods can be said to overlook certain subtleties in movement and may be less generalizable ~\cite{pmlr-v164-florence22a, codevilla2019exploring}. 

As such, we implement the implicit behavior cloning (IBC) model proposed by Florence et al. and attempt to replicate their results using a gaze behavior dataset. We use their IBC policy, MSE BC policy, and a slightly modified version of their particle environment for training and evaluation~\cite{pmlr-v164-florence22a}. Their particle environment models a moving particle aiming to reach certain setpoints while its actions are controlled via a PD controller.
The original dataset for this particle environment is artificially and randomly generated.  We want to expand on their work by using real expert demonstrations generated by a human facilitator who is leading a social interaction. 
To successfully replicate their results, we expect to see a similar model performance for both the IBC and MSE implementations based on their paper's reported success metric. To successfully extend their research, we are evaluating the naturalness of the models' generated behaviors using SPARC~\cite{SPARC_2024,SPARC_2020,SPARC_2021}, and \( R^2 \).

\section{Method}\label{sec:method}
\subsection{Problem Formulation}\label{sec:method-1-problem-graph}
We want to train our behavior cloning models using real human behavior. To do this, we must simplify the problem in two ways. Firstly, we attempt to replicate gaze behavior before full body movements to simplify our observation space\cite{eyegaze2021}. Secondly, we reduce the dimensionality of the gaze vectors to 2D space through its representation in spherical coordinates\cite{L2CS_2022}. With our problem simplified, we specify the goal as training an IBC model to generate a gaze trajectory between the facilitator's head position at time $t$ and his head position at time $t+n$. We maintain this approach when attempting to generate pose behavior.
\subsection{Multiparty Interaction Dataset}\label{sec:method-2-data}

\paragraph{Data Collection}
We want to use human expert data for training. We use a subset of the FUMI-MPF dataset, which originates from an experiment in which participants' behaviors are monitored during various large social interaction settings. Each social interaction is led by a facilitator from one of three distinct facilitator types: teacher, music teacher, or musician~\cite{roman2024}. For this study, we selected seven sessions featuring a six-participant setup including the facilitator. We also used an additional nine sessions with a four-participant setup. Independent of the number of participants, each session is comprised of about 30,000 frames. With the slightly smaller 7-session dataset, we use 2-fold validation with a 4:3 ratio for fold 1 and a 5:2 ratio for fold 2. We mainly utilize the image data from the dataset. These images were taken with a Theta Z1 camera, which provides a 360-degree view of the scene.

\paragraph{Pre-Processing}
The data is pre-processed so that it can be used with the PD controller from Florence et al.'s particle environment ~\cite{pmlr-v164-florence22a}. L2CS-Net is particularly useful for this dataset as we use it to predict the gaze of each participant ~\cite{abdelrahman2023l2cs}. The pitch and yaw of each participant's head are estimated at each frame. Actions are then determined by calculating the differences in the facilitator's pitch and yaw between consecutive frames. Each training sample is called an episode. An episode is defined as a segment of 50 frames, with the facilitator's gaze at the 50th frame set as the final setpoint we want to achieve during an episode. Observations are represented by the gaze directions of all participants at each frame. 
We represent the gaze vectors as $\textit{G}=\{\text{yaw}, \text{pitch}\}$. Since the Theta Z1 camera records at 30FPS, we can find the change in time between two frames as $1/30s$. We divide the change in gaze by the change in time to get gaze velocity at each frame. The first frame of each video is assumed to have a velocity of 0 $rad/s$. For the rest of the paper, we will refer to the processed data using the definitions below:
\begin{enumerate} \label{features}
    
    \item The facilitator's head position at time $\textit{t}$ is denoted as $\textit{G}^{t}_{f}$
    
    \item The rate of change of the facilitator's head position $\Delta \textit{G}_{f}/{\Delta t}$
    
    \item The final target position of the facilitator's head
    $\textit{G}^{t+n}_{f}$, where $n$ is the length of each episode
    
    \item The head position of the $\textit{i}^{th}$ participant at time $\textit{t}$ is denoted as $\textit{G}^{t}_{i}$, where $1 \leq i \leq 5$
    
\end{enumerate}

\subsection{Model}\label{model information}
\begin{figure}[ht!] 
  \centering
 \includegraphics[width=0.40\textwidth]{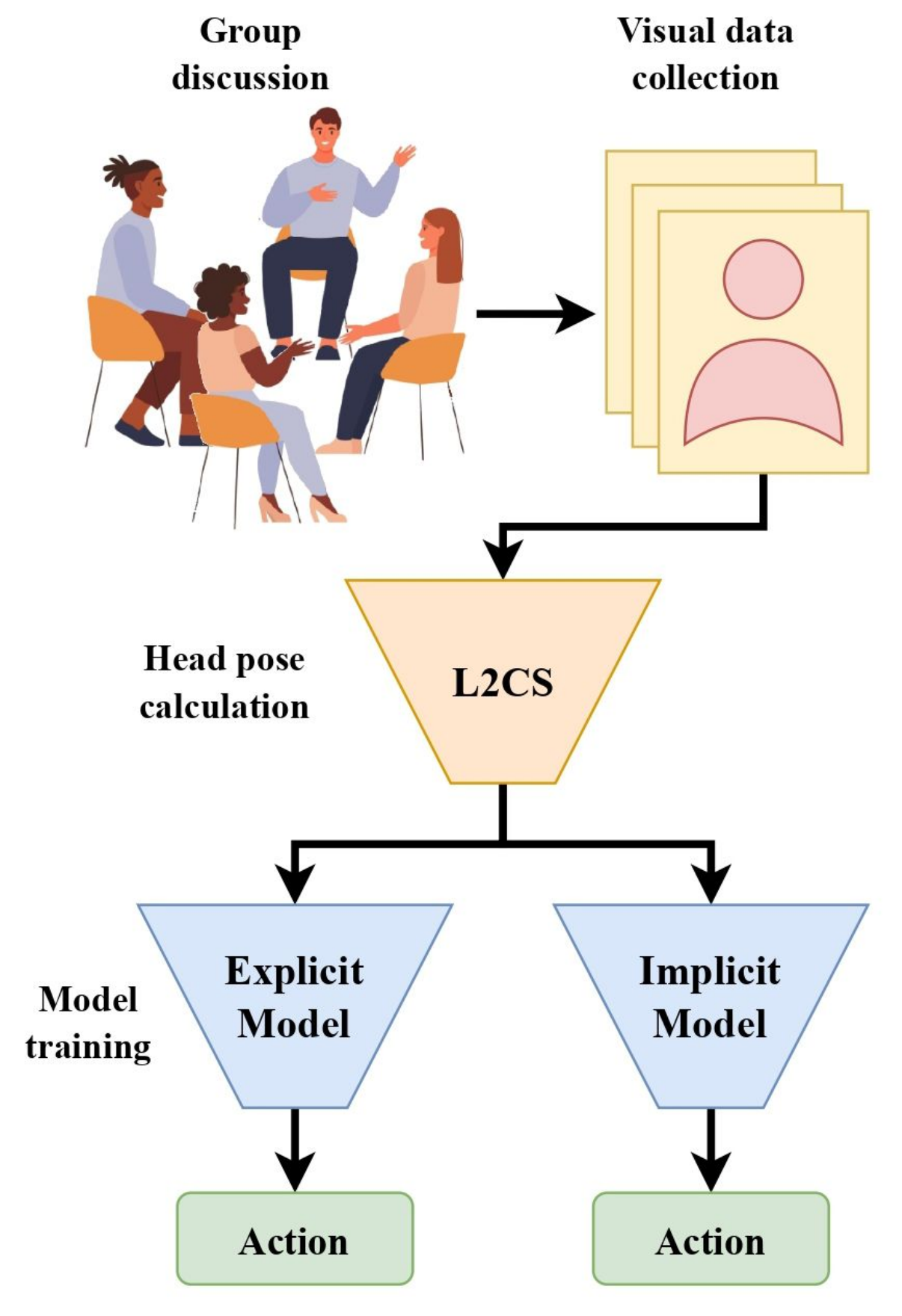}
  \caption{An overview of the pipeline for training both models.}
  \label{fig:data_struct}
\end{figure}

\subsubsection{Collecting Expert Demonstrations}
To train a behavior cloning model, we need to represent our data as a set of action and observation pairs\cite{bcorigin}. 

\begin{equation}
 D =\{ (o_1, a_1), (o_2, a_2)...  (o_n, a_n)\}
\end{equation}

The observations and actions are described in \ref{features}. The expert actions are represented by the change in the facilitator's head position. The start of an episode is the environment at time $t$. We then sample the next 49 frames from our dataset and set the episode's goal setpoint as $t+49$. Then, for each step in the episode, we read the outcomes of the previous action from the environment and use those observations along with the previous action to generate the next action. 

\subsubsection{Model the Policy}
For both the explicit and implicit behavior cloning models, we use the same multi-layer perceptron (MLP) for generating an action. The MLP is composed of a normalization layer, followed by an activation layer, which is then followed by a dropout layer, and then completed by a dense layer along with a final de-normalization layer~\cite{pmlr-v164-florence22a}. While the MLP is the same, both models still generate actions differently.

\textbf{IBC Policy}\label{ibc-policy}:
 At inference time, our initial actions are sampled from a uniform distribution modeled on our expert data. These actions are passed, along with our scene observations, through an MLP. This MLP represents our energy-based model (EBM), $E_\theta(o, a)$. We then use the calculated gradients of the EBM for our action sample combined with the Langevin MCMC (Markov Chain Monte Carlo) sampling technique from~\cite{pmlr-v164-florence22a, NEURIPS2019_378a063b} to refine our generated action. We pass the new low-energy action to a controller. For a more detailed description of EBMs and the Langevin MCMC sampling technique please refer to the work done in~\cite{pmlr-v164-florence22a, NEURIPS2019_378a063b}.

\textbf{MSE Policy}\label{mse-policy}:
To generate an action with the baseline MSE policy, we pass an observation through the MLP and the final logits represent the predicted action.

\subsubsection{Model the Environment}
Our environment is based on the particle environment described in Appendix C.5 in~\cite{pmlr-v164-florence22a}. 
The environment is structured around controlling a particle with a PD controller, which receives actions from the policy as setpoints for the particle to achieve. In the original study, the particle environment features two goals per episode, with the particle transitioning to the second goal after reaching the first. For our replication study, training and inference conclude after reaching the first goal. Additionally, we focus on tracking the facilitator's gaze behavior instead of the particle's dynamics.

\subsection{Training}
\subsubsection{IBC Training}\label{ibc-training}
Our IBC training follows the description in Appendix B.3 of~\cite{pmlr-v164-florence22a}. As referenced in \ref{ibc-policy}, we create a uniform distribution using the minimum and maximum action from our dataset. The full range of the environment is different from the full range in the dataset. Meaning, that our environment has an action space of $\{\frac{-\pi}{2}$,$\frac{\pi}{2}\}$ but is clipped to $\{a_{max}, a_{min}\}$. We then generate negative sample actions by sampling from our uniform distribution. For each negative sample, we compute its energy using the EBM:
\[
E_\theta(o, \bar{a}_i) \text{ where } E_i = E_\theta(o, \bar{a}_i)
\]
For each negative sample \( a_i^{-} \), we perform the following iterative updates for the steps $N_{MCMC}$.

\textbf{Compute Gradient}:
Calculate the gradient of the energy with respect to the action:
\[
\nabla_a E_i = \nabla_a E_\theta(o, \bar{a}_i)
\]
where \(\nabla_a E_i\) is the gradient of the EBM with respect to the action \(a_i\) and \(E_\theta\) is the EBM parameterized by \(\theta\).

\textbf{Update the Action}:
Use the Langevin MCMC update rule to refine the action:
\[
a_{i, \text{new}}^{-} = a_i^{-} - \eta \nabla_a E_i + \epsilon
\]
where \(a_{i, \text{new}}^{-}\) is the updated action, \(a_i^{-}\) is the current action, and \(\eta\) is the step size, \(\epsilon\) is a noise term..

Once completed, we soft-max the new negative energies and the energy assigned to the ground truth. We then compare them to a binary ground truth array with the true action as 1 and the negative actions as 0. We compute the loss between the two arrays using an InfoNCE loss function\cite{InfoNCE2018}. For back-propagation, we use an Adam optimizer as described in Appendix B \cite{pmlr-v164-florence22a}.

\textbf{Hyper-parameters}: We can find a list of the parameters used for training in Appendix D.3 of~\cite{pmlr-v164-florence22a}. However, the paper's public repository\footnote{https://github.com/google-research/ibc} with the implemented solutions has a separate set of recommended hyper-parameters. For this replication study, we use the latter set of hyper-parameters.

\subsubsection{MSE Training}
The explicit behavior cloning model is trained using the MSE loss between the ground truth action and the predicted action from the MLP. The predicted actions are still clipped to $\{a_{max}, a_{min}\}$. We use the same Adam Optimizer for our explicit model back-propagation as for our implicit model.

\textbf{Hyper-parameters}: As stated previously, we use the hyper-parameters laid out in the publicly available implementation of the work done in~\cite{pmlr-v164-florence22a}.
\subsection{Evaluation}
\subsubsection{Inference}
This section contains a description of the inference process for each model. The main difference between the models is that the IBC model uses actions and observations to perform inference whereas the MSE model uses solely observations.

\textbf{IBC Inference}:
We sample from the uniform distribution created in Section \ref{ibc-training} to generate an initial action \( a_0 \). This action is passed along with the observation \( o_0 \) through our EBM. We compute the gradient:
\[
\nabla_a E = \nabla_a E_\theta(o_0, a_0)
\]
We then apply the Langevin MCMC technique described in Sections \ref{ibc-policy} and \ref{ibc-training}. After we run the process for \( N_{MCMC} \) steps we take the final action and then pass it through a second time.

\textbf{MSE Inference}:
Given an observation \( o \), the MLP computes the predicted action \( \hat{a} \) as follows:
\[
    \hat{a} = F_{\theta}(o)
\]

\subsubsection{Evaluation Metrics}
We utilize the average success metric (ASM) from ~\cite{pmlr-v164-florence22a}, which assigns a value of 1.0 or 0.0 to episodes based on whether the policy successfully reaches an episode's goal setpoint. In the particle environment if the Euclidean distance between the target and current positions is within a threshold of 0.02 radians ~\cite{pmlr-v164-florence22a}, the episode scores 1.0. We then average these scores to obtain the ASM. We also use the $R^2$ of the predicted trajectories to assess how well the IBC model replicates the ground truth human behavior. Finally, to test the smoothness of the generated behavior, we use the SPARC metric which is commonly used to evaluate human movement \cite{SPARC_2020, SPARC_2021, SPARC_2024}. In our implementation the convention is that values closer to 0.0 are smoother.
\section{Results \& Discussion}\label{sec:results}
\subsubsection{AverageSuccessMetric}Florence et al. claim that they are "able to train $95\%$ successful implicit policies up to 16 dimensions, whereas explicit (MSE) policies can only do 8 dimensions with the same success rate"~\cite{pmlr-v164-florence22a}.\begin{table}[ht]
\normalsize
\centering
\begin{tabular}{lrrr}
\hline
\textbf{Facilitator} & \textbf{Implicit BC} & \textbf{Explicit BC}  & \textbf{Reported} \\ \hline
Teacher                   & 0.96                 & 0.92                           & 0.95              \\
Musician                & \textbf{0.97}                 & \textbf{0.95}                           & 0.95              \\
M. Teacher            & 0.96                 & 0.93                          & 0.95              \\ \hline
Average       & \textbf{0.96}        & 0.93    & 0.95     \\ \hline
\end{tabular}
\caption{Model performance comparison using average success metric where 1.0 is the highest success rate and 0.0 is the lowest.}
\end{table}
\label{asm-results}
The results in \ref{asm-results} show that the average success rate averaged out over the sessions is at $96\%$ for the IBC model. The explicit model, however, is at $93\%$ which is slightly below the reported $95\%$. 
We see that the best dataset for both models is the session with a musician as the facilitator.
\subsubsection{\texorpdfstring{$R^2$}{R2}} In table \ref{r2-results} we see that the best average $R^2$ value for both pitch and yaw is achieved by the MSE model at $76\%$ for average pitch and $56\%$ for average yaw. Similar to the ASM the Musician has the best results at $67\%$ for IBC and $75\%$ for MSE but only in yaw. The best results for pitch occur for the session using a music teacher at $83\%$ and $86\%$.
\begin{table}[ht]
\normalsize
\centering
\begin{tabular}{l|rr|rr}
\hline
$\mathbf{R^2}$       & \multicolumn{2}{c|}{\textbf{Implicit BC}} & \multicolumn{2}{c}{\textbf{Explicit BC}} \\ 
                      & \textbf{Pitch} & \textbf{Yaw}             & \textbf{Pitch} & \textbf{Yaw}             \\ \hline
Teacher               & 0.79           & 0.13                    & 0.81           & 0.35                      \\ 
Musician              & 0.57           & \textbf{0.67}           & 0.61           & \textbf{0.75}             \\ 
M. Teacher            & \textbf{0.83}  & 0.44                    & \textbf{0.86}  & 0.58                      \\ \hline
Average               & 0.73           & 0.41                    & \textbf{0.76}  & \textbf{0.56}             \\ \hline
\end{tabular}
\caption{Model performance comparison using $R^2$ as a comparison metric where 1.0 is the best fit and 0.0 is the worst fit.}
\end{table}\label{r2-results}
\subsubsection{SPARC Metric}
In table \ref{sparc-results} we see that the musician session is the best performing for IBC and that the music teacher has the best result in pitch for MSE with the teacher session having the best result in yaw. However, the musician session performs quite well overall. Looking at the overall average, we see that IBC has a better performance.
\begin{table}[ht]
\normalsize
\centering
\begin{tabular}{l|rr|rr}
\hline
\textbf{SPARC Metric} & \multicolumn{2}{c|}{\textbf{Implicit BC}} & \multicolumn{2}{c}{\textbf{Explicit BC}} \\ 
                      & \textbf{Pitch} & \textbf{Yaw}             & \textbf{Pitch} & \textbf{Yaw}             \\ \hline
Teacher               & -49.91         & -276.60                 & -244.24        & \textbf{-404.91}         \\ 
Musician              & \textbf{-29.72} & \textbf{-178.05}         & -203.50        & -410.86                  \\ 
M. Teacher            & -35.09         & -205.13                 & \textbf{-191.80} & -435.73                  \\\hline
Average               & \textbf{-38.02} & \textbf{-220.82}         & -213.85        & -417.16                  \\ \hline
\end{tabular}
\caption{Combined averages for SPARC Pitch and SPARC Yaw values. Values closer to zero are smoother signals.}
\end{table}\label{sparc-results}

The ASM confirms our successful replication of the findings from Florence et al.~\cite{pmlr-v164-florence22a}. Our EBM achieved an average success rate of $96\%$, closely aligning with the $95\%$ reported in their study. In contrast, the MSE model fell short, recording an overall average success rate of only $93\%$. As illustrated in Figure 5 of Florence et al., this performance gap is not uncommon, suggesting that the MSE model may be less effective in certain scenarios.
Interestingly, the $R^2$ results present a different perspective, indicating that the MSE model better replicates the trajectory of our facilitators. Specifically, the average $R^2$ for pitch and yaw in the MSE model surpassed those of the IBC model by $3\%$ and $15\%$, respectively.

The SPARC metrics are averaged for each facilitator type across all sessions. The IBC results have an $82\%$ increase in pitch and a $47\%$ increase in yaw performance than the explicit model. Since smoother motion indicates normally functioning joints \cite{SPARC_2020, SPARC_2021, SPARC_2024}, it is plausible that the IBC model generates more natural motions. 
From all three metrics, we surmise that the IBC predictions are more likely to attain the correct final gaze position but over a different trajectory than the ground truth. Furthermore, the IBC trajectories are smoother and more natural than the MSE ones according to the SPARC metric. Looking at the facilitator types, the musician facilitator seems to consistently perform better, probably due to the model's being trained on a small custom dataset.  %
We also attempted to expand our implementation to pose generation but faced challenges. To achieve usable results, we had to set an unreasonably high MSE threshold of 500.0 pixels for the average success metric, and the generated keypoints lacked humanoid qualities. This issue stemmed from PD controller overshoot, which distorted the keypoints, and the increased action space complexity, which negatively impacted both models. Future work could explore per-joint models or alternatives to PD controllers.
Despite these challenges, our results suggest that implicit behavior cloning can generate more natural, human-like behaviors than traditional methods, offering promising directions for future research.

\section{Conclusion \& Future Works}

This study successfully replicated the IBC model to generate nonverbal cues like gaze behavior for multiparty facilitation, which is supported by consistent results across multiple commonly used metrics. However, due to the lack of suitable datasets for different social interaction tasks, this study has not investigated the reason of the musician facilitator consistently performed better across sessions. Future research will explore what makes certain behaviors easier to learn. Another limitation is the absence of human evaluators. In the next stage, we plan to implement these models in live social interactions, such as teaching scenarios and social games, to evaluate the model's generalization performance in real-world settings.

\section*{ACKNOWLEDGMENT}
 This research is the result of a tripartite research collaboration between the University of Tokyo, Hiroo Gakuen Junior Senior High School, and Honda Research Institute Japan. The authors thank the collaborators for their assistance in implementing the experiments.

\bibliographystyle{IEEEtran}
\bibliography{bib}

\begin{thebibliography}{10}
\providecommand{\url}[1]{#1}
\csname url@samestyle\endcsname
\providecommand{\newblock}{\relax}
\providecommand{\bibinfo}[2]{#2}
\providecommand{\BIBentrySTDinterwordspacing}{\spaceskip=0pt\relax}
\providecommand{\BIBentryALTinterwordstretchfactor}{4}
\providecommand{\BIBentryALTinterwordspacing}{\spaceskip=\fontdimen2\font plus
\BIBentryALTinterwordstretchfactor\fontdimen3\font minus \fontdimen4\font\relax}
\providecommand{\BIBforeignlanguage}[2]{{%
\expandafter\ifx\csname l@#1\endcsname\relax
\typeout{** WARNING: IEEEtran.bst: No hyphenation pattern has been}%
\typeout{** loaded for the language `#1'. Using the pattern for}%
\typeout{** the default language instead.}%
\else
\language=\csname l@#1\endcsname
\fi
#2}}
\providecommand{\BIBdecl}{\relax}
\BIBdecl

\bibitem{chen2024artificial}
J.~J. Chen and J.~C. Lin, ``Artificial intelligence as a double-edged sword: Wielding the power principles to maximize its positive effects and minimize its negative effects,'' \emph{Contemporary Issues in Early Childhood}, vol.~25, no.~1, pp. 146--153, 2024.

\bibitem{hickok2024public}
M.~Hickok, ``Public procurement of artificial intelligence systems: new risks and future proofing,'' \emph{AI \& society}, vol.~39, no.~3, pp. 1213--1227, 2024.

\bibitem{chiang2024enhancing}
C.-W. Chiang, Z.~Lu, Z.~Li, and M.~Yin, ``Enhancing ai-assisted group decision making through llm-powered devil's advocate,'' in \emph{Proceedings of the 29th International Conference on Intelligent User Interfaces}, 2024, pp. 103--119.

\bibitem{zhang2022storybuddy}
Z.~Zhang, Y.~Xu, Y.~Wang, B.~Yao, D.~Ritchie, T.~Wu, M.~Yu, D.~Wang, and T.~J.-J. Li, ``Storybuddy: A human-ai collaborative chatbot for parent-child interactive storytelling with flexible parental involvement,'' in \emph{Proceedings of the 2022 CHI Conference on Human Factors in Computing Systems}, 2022, pp. 1--21.

\bibitem{liu2024peergpt}
J.~Liu, Y.~Yao, P.~An, and Q.~Wang, ``Peergpt: Probing the roles of llm-based peer agents as team moderators and participants in children's collaborative learning,'' in \emph{Extended Abstracts of the CHI Conference on Human Factors in Computing Systems}, 2024, pp. 1--6.

\bibitem{agent2021agent}
F.~Agent, ``An agent that facilitates crowd discussion,'' 2021.

\bibitem{phutela2015importance}
D.~Phutela, ``The importance of non-verbal communication,'' \emph{IUP Journal of Soft Skills}, vol.~9, no.~4, p.~43, 2015.

\bibitem{eyegaze2021}
C.~Ishi and T.~Shintani, ``Analysis of eye gaze reasons and gaze aversions during three-party conversations,'' in \emph{Interspeech 2021}, 08 2021, pp. 1972--1976.

\bibitem{gillet2022learning}
S.~Gillet, M.~T. Parreira, M.~V{\'a}zquez, and I.~Leite, ``Learning gaze behaviors for balancing participation in group human-robot interactions,'' in \emph{2022 17th ACM/IEEE International Conference on Human-Robot Interaction (HRI)}.\hskip 1em plus 0.5em minus 0.4em\relax IEEE, 2022, pp. 265--274.

\bibitem{rosenberg2020}
R.~Rosenberg-Kima, Y.~Koren, and G.~Gordon, ``Robot-supported collaborative learning (rscl): Social robots as teaching assistants for higher education small group facilitation,'' \emph{Frontiers in Robotics and AI}, vol.~6, 01 2020.

\bibitem{pmlr-v164-florence22a}
\BIBentryALTinterwordspacing
P.~Florence, C.~Lynch, A.~Zeng, O.~A. Ramirez, A.~Wahid, L.~Downs, A.~Wong, J.~Lee, I.~Mordatch, and J.~Tompson, ``Implicit behavioral cloning,'' in \emph{Proceedings of the 5th Conference on Robot Learning}, ser. Proceedings of Machine Learning Research, A.~Faust, D.~Hsu, and G.~Neumann, Eds., vol. 164.\hskip 1em plus 0.5em minus 0.4em\relax PMLR, 08--11 Nov 2022, pp. 158--168. [Online]. Available: \url{https://proceedings.mlr.press/v164/florence22a.html}
\BIBentrySTDinterwordspacing

\bibitem{codevilla2019exploring}
F.~Codevilla, E.~Santana, A.~M. L{\'o}pez, and A.~Gaidon, ``Exploring the limitations of behavior cloning for autonomous driving,'' in \emph{Proceedings of the IEEE/CVF international conference on computer vision}, 2019, pp. 9329--9338.

\bibitem{SPARC_2024}
\BIBentryALTinterwordspacing
G.~Cornec, M.~Lempereur, J.~Mensah-Gourmel, J.~Robertson, L.~Miramand, B.~Medee, S.~Bellaiche, R.~Gross, J.-M. Gracies, O.~Remy-Neris, and N.~Bayle, ``Measurement properties of movement smoothness metrics for upper limb reaching movements in people with moderate to severe subacute stroke,'' \emph{Journal of NeuroEngineering and Rehabilitation}, vol.~21, no.~1, May 2024. [Online]. Available: \url{https://pubmed.ncbi.nlm.nih.gov/38812037/}
\BIBentrySTDinterwordspacing

\bibitem{SPARC_2020}
\BIBentryALTinterwordspacing
A.~I. Figueiredo, G.~Balbinot, F.~O. Brauner, A.~Schiavo, R.~R. Baptista, A.~S. Pagnussat, K.~Hollands, and R.~G. Mestriner, ``Sparc metrics provide mobility smoothness assessment in oldest-old with and without a history of falls: a case control study,'' \emph{Frontiers in Physiology}, vol.~11, Jun. 2020. [Online]. Available: \url{https://pmc.ncbi.nlm.nih.gov/articles/PMC7298141/}
\BIBentrySTDinterwordspacing

\bibitem{SPARC_2021}
\BIBentryALTinterwordspacing
M.~Saes, M.~I.~M. Refai, J.~Van~Kordelaar, B.~L. Scheltinga, B.-J.~F. Van~Beijnum, J.~B.~J. Bussmann, J.~H. Buurke, P.~H. Veltink, C.~G.~M. Meskers, E.~E.~H. Van~Wegen, and G.~Kwakkel, ``Smoothness metric during reach-to-grasp after stroke: part 2. longitudinal association with motor impairment,'' \emph{Journal of NeuroEngineering and Rehabilitation}, vol.~18, no.~1, Sep. 2021. [Online]. Available: \url{https://doi.org/10.1186/s12984-021-00937-w}
\BIBentrySTDinterwordspacing

\bibitem{L2CS_2022}
\BIBentryALTinterwordspacing
A.~A. Abdelrahman, T.~Hempel, A.~Khalifa, and A.~Al-Hamadi, ``L2cs-net: Fine-grained gaze estimation in unconstrained environments,'' \emph{arXiv (Cornell University)}, Jan. 2022. [Online]. Available: \url{https://arxiv.org/abs/2203.03339}
\BIBentrySTDinterwordspacing

\bibitem{roman2024}
\BIBentryALTinterwordspacing
J.~Y. Chew and K.~Nakamura, ``Who to teach a robot to facilitate multi-party social interactions?'' in \emph{Companion of the 2023 ACM/IEEE International Conference on Human-Robot Interaction}, ser. HRI '23.\hskip 1em plus 0.5em minus 0.4em\relax New York, NY, USA: Association for Computing Machinery, 2023, p. 127–131. [Online]. Available: \url{https://doi.org/10.1145/3568294.3580056}
\BIBentrySTDinterwordspacing

\bibitem{abdelrahman2023l2cs}
A.~A. Abdelrahman, T.~Hempel, A.~Khalifa, A.~Al-Hamadi, and L.~Dinges, ``L2cs-net: Fine-grained gaze estimation in unconstrained environments,'' in \emph{2023 8th International Conference on Frontiers of Signal Processing (ICFSP)}.\hskip 1em plus 0.5em minus 0.4em\relax IEEE, 2023, pp. 98--102.

\bibitem{bcorigin}
\BIBentryALTinterwordspacing
S.~Schaal, ``Learning from demonstration,'' in \emph{Advances in Neural Information Processing Systems}, M.~Mozer, M.~Jordan, and T.~Petsche, Eds., vol.~9.\hskip 1em plus 0.5em minus 0.4em\relax MIT Press, 1996. [Online]. Available: \url{https://proceedings.neurips.cc/paper_files/paper/1996/file/68d13cf26c4b4f4f932e3eff990093ba-Paper.pdf}
\BIBentrySTDinterwordspacing

\bibitem{NEURIPS2019_378a063b}
\BIBentryALTinterwordspacing
Y.~Du and I.~Mordatch, ``Implicit generation and modeling with energy based models,'' in \emph{Advances in Neural Information Processing Systems}, H.~Wallach, H.~Larochelle, A.~Beygelzimer, F.~d\textquotesingle Alch\'{e}-Buc, E.~Fox, and R.~Garnett, Eds., vol.~32.\hskip 1em plus 0.5em minus 0.4em\relax Curran Associates, Inc., 2019. [Online]. Available: \url{https://proceedings.neurips.cc/paper_files/paper/2019/file/378a063b8fdb1db941e34f4bde584c7d-Paper.pdf}
\BIBentrySTDinterwordspacing

\bibitem{InfoNCE2018}
\BIBentryALTinterwordspacing
A.~Van Den~Oord, Y.~Li, and O.~Vinyals, ``\BIBforeignlanguage{en}{Representation learning with contrastive predictive coding},'' Jul. 2018. [Online]. Available: \url{https://arxiv.org/abs/1807.03748}
\BIBentrySTDinterwordspacing

\end{thebibliography}

\end{document}